# M2H: Multi-Task Learning with Efficient Window-Based Cross-Task Attention for Monocular Spatial Perception

U.V.B.L Udugama*, George Vosselman, Francesco Nex

*Abstract*— Deploying real-time spatial perception on edge devices requires efficient multi-task models that leverage complementary task information while minimizing computational overhead. In this paper, we introduce Multi-Mono-Hydra (M2H), a novel multi-task learning framework designed for semantic segmentation and depth, edge, and surface normal estimation from a single monocular image. Unlike conventional approaches that rely on independent single-task models or shared encoder-decoder architectures, M2H introduces a Window-Based Cross-Task Attention Module that enables structured feature exchange while preserving task-specific details, improving prediction consistency across tasks. Built on a lightweight ViT-based DINOv2 backbone, M2H is optimized for real-time deployment and serves as the backbone for monocular spatial perception systems, a framework for 3D scene graph construction in dynamic environments. Comprehensive evaluations demonstrate that M2H outperforms state-of-the-art (SOTA) multi-task models on NYUDv2, exceeds single-task depth and semantic baselines on Hypersim, and achieves superior performance on Cityscapes datasets, all while maintaining computational efficiency on laptop hardware. Beyond curated benchmarks, we validate M2H on real-world data, demonstrating its practicality in spatial perception tasks. We provide our implementation and pretrained models at *https://github.com/UAV-Centre-ITC/M2H.git*.

## I. INTRODUCTION

Dense vision-based scene understanding lies at the heart of autonomous systems, augmented reality, and robotic perception. By jointly analyzing key pixel-level tasks, such as semantic segmentation and depth, edge, and surface normal estimation, multi-task learning (MTL) offers a scalable approach to real-time decision-making in complex environments. Drawing on seminal insights from Taskonomy [1], where inter-task relationships were shown to dramatically reduce supervision requirements and improve generalization, MTL frameworks have made it feasible to capture complementary cues across tasks. For instance, depth discontinuities often align with semantic boundaries, while surface normals share information with both depth estimation and object segmentation. However, effectively exploiting these cross-task synergies remains non-trivial.

Some approaches operate with localized or final-stage interactions, preserving efficiency but limiting deeper synergy, while others employ full global attention for richer context at the cost of high computational overhead. Consequently, there is a growing need for an MTL framework that balances continuous cross-task feature sharing with practical efficiency.

In this work, we propose M2H, a real-time MTL framework employing a lightweight ViT-based backbone and a window-based strategy to capture and exchange both local and global cues. Our model significantly outperforms prior multi-task methods on the NYUDv2 dataset, achieving a +3.4% mIoU improvement in semantic segmentation and a 13% lower RMSE in depth estimation. On Hypersim, where no prior multi-task benchmarks exist, M2H surpasses the single-task depth model Scale Depth-NK [2] with a 33% lower RMSE, and in semantic segmentation, it achieves +5.4 mIoU over EMSANet [3], a SOTA segmentation model. Additionally, M2H operates at 30 FPS on an RTX 3080 laptop GPU, demonstrating its real-time feasibility. Finally, when integrated with Mono-Hydra [4], M2H enables robust 3D scene graph construction, further validating its effectiveness in real-world applications.

## II. RELATED WORK

### A. Multi-task Learning

MTL unifies multiple tasks within a single model to share representations and reduce redundant computation. Caruana [5] first introduced hard parameter sharing, branching into task-specific heads from a common trunk, a concept extended by UberNet [6] to low, mid, and high-level tasks. Large-scale analyses such as Taskonomy [1] reveal that identifying inter-task correlations reduces labeled data needs, but naive splitting can provoke negative transfer, prompting solutions like GradNorm [7] and homoscedastic uncertainty weighting [8]. Another approach, soft parameter sharing [9], merges features across distinct subnetworks, though it can add overhead. More recent MTL architectures vary considerably: state-space model decoders (e.g., MTMamba [10], MTMamba++ [11]) focus on long-range dependencies, adversarial training-based SwinMTL [12] refines each task with separate MLP heads while preserving a shared encoder-decoder, and CNN-based distillation modules (PAD-Net [13], MTI-Net [14]) unify Multimodal cues. Some frameworks optimize efficiency by adopting network binarization (e.g., Bi-MTPD [15]) to compress resource-intensive predictors and can even surpass full-precision models like ARTC [16] or InvPT [14]. Others leverage a mixture of experts for dynamic capacity allocation [17]. Despite these diverse strategies, balancing synergy and computational feasibility remains a common challenge, prompting increased attention to attention-based methods, which we discuss next.

### B. MTL with Attention-Based Modules

Among these approaches, attention mechanisms have gained traction for selectively highlighting relevant features in each task, potentially leading to richer cross-task synergy. PAD-Net [13] pioneered a multi-modal distillation module where attention guides how each target task absorbs complementary features. MTI-Net [14] further introduced multi-scale refinement to propagate task-specific cues at different network depths. ATRC [16] leverages both local and global relational contexts, enabling more expressive gating conditioned on pairwise similarities between source and target tasks.

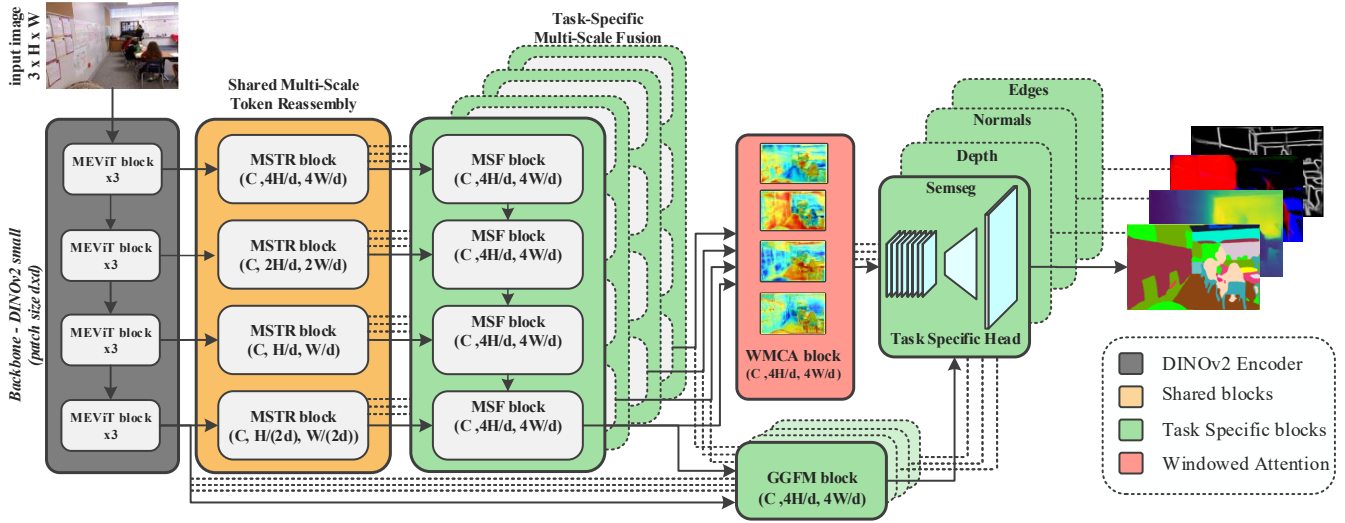

Fig. 1: Network architecture: Overview of our proposed multi-task dense prediction framework. A DINOv2 [38] small backbone first extracts multi-scale token representations using Memory Efficient ViT blocks (MEViT), which are transformed into spatial feature maps via Multi Scale Token Reassembly (MSTR) blocks. These feature maps feed into Multi Scale Fusion (MSF) blocks to produce preliminary task-specific features. A Windowed Multi-Task Cross-Attention (WMCA) block then captures local cross-task interactions within small, non-overlapping windows, while a Global Gated Feature Merging (GGFM) block aggregates global context. Finally, specialized decoder heads generate high-quality dense predictions for each task.

Recently, InvPT [18] and DenseMTL [19] introduced global attention in transformer-like decoders to unify multi-task feature representations, achieving strong performance at the cost of substantial computational overhead. While this large-scale attention can capture broad contextual cues, it often challenges real-time deployment.

### III. METHOD

#### A. Model Architecture

We propose a multi-task dense prediction framework built upon a DINOv2 [20] backbone. At a high level, it produces multi-scale token representations, which are then converted into spatial feature maps and further fused to create preliminary task-specific features. Next, a dual-path refinement strategy integrates both local cross-task interactions via windowed attention and global context via a learned gating mechanism. Finally, specialized decoder heads produce the edges, normals, semantic segmentation, and depth outputs. Fig. 1 provides a detailed, block-by-block illustration of the entire pipeline.

We now formalize the main steps of our approach: Let $x$ denote the input image. First, the DINOv2 [20] encoder produces multi-scale token representations $T$ as shown in (1).

$$T = \text{DINOv2}_{small}(x), T \in \mathbb{R}^{N \times emb_d}, \quad (1)$$

where $N$ is the number of tokens and $emb_d$ is the token dimension (in DINOv2 small patch size (d) = 16, $emb_d$ = 384, $N$ depends on the input image (Width/d*Height/d)).

These tokens are then reassembled into spatial feature maps $\{F_i\}_{i=1}^K$ via Multi-Scale Token Reassembly (MSTR) blocks as in (2), following a design similar to that in the DPT architecture [21].

$$\{F_i\}_{i=1}^K = MSTR(T), F_i \in \mathbb{R}^{B \times C \times H_i \times W_i}, \quad (2)$$

where $K$ denotes the number of scales (set to $K = 4$ in our implementation and $C=256$), the spatial dimension $H_i \times W_i$ is different at each scale as in Fig. 1.

Lightweight convolutions refine these maps, and Multi-Scale Fusion (MSF) blocks generate preliminary task-specific features $F^t$ for each dense prediction task as in (3).

$$F^t = MSR_t(\{Conv(F_i)\}_{i=1}^K), \quad (3)$$

where $t \in \{edges, normals, semantics, depth\}$.

Next, each $F^t$ is processed along two parallel paths. The cross-task local context is captured and exchanged using Windowed Multi-Task Cross-Attention (WMCA) (4a), while the task-specific global context is aggregated via Global Gated Feature Merging (GGFM) with a residual connection to the original DINOv2 [20] feature tokens $U_t$ (4b).

$$F_{local}^{\hat{t}} = WMCA(F^t) \quad (4a)$$

$$F_{global}^{\widehat{U_t+t}} = GGFM(F^t, U_t) \quad (4b)$$

The enriched features, fused by concatenating the local (WMCA) and global (GGFM) streams via a 1×1 convolution (5a), are decoded by dedicated heads to produce the final predictions $y^t$ as in (5b).

$$F_{fused}^{\hat{t}} = Conv_{1 \times 1}\left(Concat\left(F_{local}^{\hat{t}}, F_{global}^{\widehat{U_t+t}}\right)\right) \quad (5a)$$

$$y^t = Head_t(F_{fused}^{\hat{t}}) \quad (5b)$$

In the following sections, we delve into the design details of the WMCA and GGFM blocks.

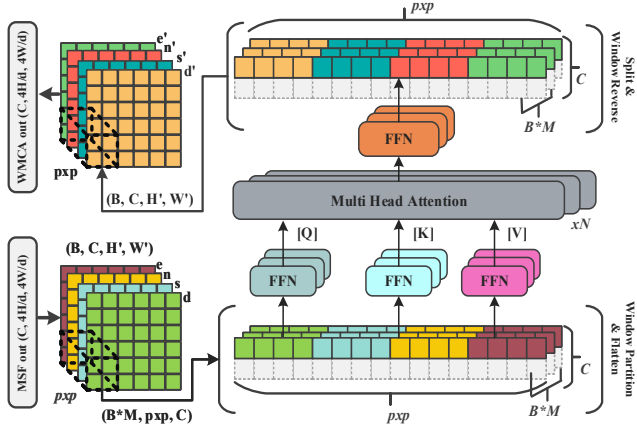

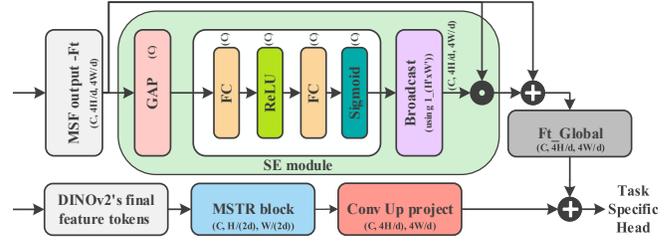

Fig. 3 : Global Context Fusion via the GGFM Block. The block receives two inputs: the task-specific feature map $F^t$ and a set of unique features $U_t$ from the DINOv2 final layer. The task-specific map is globally pooled and processed through a Squeeze-and-Excitation (SE) [39] module to generate a gating vector, which is then broadcast and used to modulate the feature map. A residual connection fuses this modulated output with the DINOv2 features, enhancing the global context of the feature flow.

Fig. 2: The WMCA block. Four task-specific feature maps $e, n, s, d$ with shape $B \times C \times H' \times W'$ (where $H' = \frac{4H}{d}, W' = \frac{4W}{d}, d = 16$, W, H are the input image dimensions), are first partitioned into $p \times p$ windows. LayerNorm is then applied to each window before the windows are flattened into tokens $B \cdot M \times 4p^2 \times C$. The normalized tokens from all tasks are concatenated and processed by a multi-head attention module (with separate Q, K, and V projections) and a feed-forward neural network (FFN), both wrapped with residual connections. Finally, the tokens are split and reshaped to reconstruct the enriched feature maps ($e'$, $n'$, $s'$, $d'$) at the original spatial resolution.

*1) Windowed Multi-Task Cross-Attention block (WMCA)*

The WMCA module is applied at the network's end, just before the task-specific heads, to efficiently exchange complementary information among task-specific features within non-overlapping local windows. Early in the decoder, shared MSTR blocks enable the sharing of initial global features. At the final stage, WMCA refines these representations by allowing tasks to exchange relevant local details. Inspired by the windowed attention mechanism of the Swin Transformer [22], our implementation is tailored for multi-task learning, ensuring efficient cross-task information sharing with minimal computational overhead.

In our approach, as illustrated in Fig. 2, each task-specific feature map edges $e$, surface normals $n$, semantics $s$, and depth $d$ with dimensions, $e, n, s, d \in R^{B \times C \times H' \times W'}$, is partitioned into small non-overlapping windows of size $p \times p$ (with zero padding applied if $H'$ and $W'$ are not multiples of $p$). This transformation flattens each $p \times p$ window into a $p^2$-length sequence of tokens as in (6). In each window, the tokens are normalized using Layer Normalization and then concatenated across tasks to form a combined representation (7).

For each $x \in \{e, n, s, d\}$,

$$x_{win} \in \mathbb{R}^{B \cdot M \times p^2 \times C}, \text{ with } M = \frac{H'W'}{p^2}. \quad (6)$$

$$Z_{in} = \begin{bmatrix} LN(e_{win}), LN(n_{win}), \\ LN(s_{win}), LN(d_{win}) \end{bmatrix} \in \mathbb{R}^{B \cdot M \times 4p^2 \times C}, \quad (7)$$

where $e_{win}, n_{win}, s_{win}, d_{win}$ are the window-partitioned versions of $e, n, s, d$ respectively.

Learned projections compute query, key, and value matrices, and multi-head cross-attention is performed so that the attention output is added to the input tokens (8).

$$Z_{attn} = Z_{in} + A, \quad A = \text{softmax}\left(\frac{QK^T}{\sqrt{d_h}}\right)V, \quad (8)$$

where $Q = Z_{in}W_Q$, $K = Z_{in}W_K$, $V = Z_{in}W_v$.

A feed-forward network refines the attended features with a residual connection as in (9).

$$Z_{ffn} = Z_{attn} + FFN(Z_{attn}), \quad (9)$$

where $FFN(z) = \sigma(zW_1 + b_1)W_2 + b_2$, $\sigma$ is a non-linear activation (e.g., GELU), and $W_1, W_2, b_1, b_2$ are learnable parameters.

Finally, the refined tokens $Z_{ffn}$, are split back into four groups and rearranged to reconstruct the spatial feature maps for each task as in (10a).

$$e'_{win}, n'_{win}, s'_{win}, d'_{win} = Z_{ffn} \quad (10a)$$

where each tensor has the shape of $\mathbb{R}^{B \cdot M \times p^2 \times C}$ and reshaped into

$$e', n', s', d' \in \mathbb{R}^{B \times C \times H' \times W'}. \quad (10b)$$

In this way, the module transforms the input feature maps $(e, n, s, d)$ into enriched outputs $(e', n', s', d')$, that incorporate local cross-task context (10b). Within the M2H decoder, we employed a WMCA block consisting of two layers, each featuring four multi-head attention heads and a window size of 7.

*2) Global Gated Feature Merging (GGFM)*

The GGFM block illustrated in Fig. 3 aggregates global context for each task-specific feature map $F^t \in \mathbb{R}^{B \times C \times H' \times W'}$ using a learned gating mechanism. In our design, we first compute a global descriptor $z^t$ from $F^t$ (3) using global average pooling (11). Next, a two-layer MLP with ReLU and sigmoid activations generates a gating vector $g^t$ (12). Finally, this gating vector is broadcast over the spatial dimensions and applied to $F^t$ via elementwise multiplication. The result is then added back to $F^t$ through a residual connection to produce the enriched feature map $F^{\hat{t}}_{global}$ (13).

$$z^t = GAP(F^t) \in \mathbb{R}^{B \times C}, \quad (11)$$

$$g^t = \sigma(W_2(ReLU(W_1 z^t)) + b) \in \mathbb{R}^{B \times C}, \quad (12)$$

$$F^{\hat{t}}_{global} = F^t + (g^t \otimes 1_{H' \times W'}) \odot F^t, \quad (13)$$

where $GAP$ denotes global average pooling, $\sigma$ is the sigmoid function, $W_1, W_2,$ and $b$ are learnable parameters, $\otimes$ denotes broadcasting of $g^t$ using a matrix of ones $1_{H' \times W'}$ to match the spatial dimensions of $F^t$, and $\odot$ is elementwise multiplication.

Additionally, the final DINOv2 token map, converted into a spatial feature map using the MSTR block and up projections $U^t$, is fused with $F_{global}^{\hat{t}}$ to further refine the global context.

$$F_{global}^{\widehat{U_t+t}} = F_{global}^{\hat{t}} + U^t \tag{14}$$

### B. Loss Functions

Our multi-task framework employs six losses: four task-specific losses, segmentation, depth, surface normals, and edges, and two cross-task consistency losses, depth–normal and edge–segmentation. These losses ensure that both individual task performance and cross-task coherence are addressed during training.

*1) Task Specific Losses*

**Segmentation Loss** is computed as a combination of Cross-Entropy and Dice losses as in (14):

$$L_{seg} = \alpha L_{ce} + \beta L_{dice}, \tag{14}$$

During initial training, we use $\alpha = 0.5$ and $\beta = 0.75$ ; during fine-tuning, we adjust to $\alpha = 0.75$ and $\beta = 1.0$ .

**Depth loss** is tailored to the dataset. For indoor scenes, the loss is defined as (15a):

$$L_{depth} = L_{huber} + \lambda_{grad} \parallel \nabla p - \nabla y \parallel^2 \tag{15a}$$

where $L_{huber}$ is the Huber loss between the predicted depth $p$ and the ground truth $y$.

For outdoor scenes, to better handle the wide range of depth values, we replace the Huber loss with a scale-invariant loss, and compute the gradient loss on the logarithm of the depth values (15b):

$$L_{depth} = L_{si} + \lambda_{grad} \parallel \nabla(\log p) - \nabla(\log y) \parallel^2. \tag{15b}$$

We set $\lambda_{grad}$ to values ranging from 0.5 to 2, which control the contribution of the gradient loss.

**Surface Normals** $L_{normals}$ Loss uses cosine similarity to quantify the angular difference between the predicted normals and the true normals.

**The edge loss** $L_{edges}$ employs binary cross-entropy with logits to evaluate the network's performance in detecting image boundaries.

*2) Cross-Task Consistency Losses*

**Depth-Normal Consistency Loss**; given a predicted depth map $d$, we approximate the local depth gradients using finite differences: $\partial_x d = d(x, y + 1) - d(x, y)$ and $\partial_y d = d(x + 1, y) - d(x, y)$. These approximations yield the corresponding normals as in (16a):

$$n_{depth} = \frac{|-\partial_x d, -\partial_y d, 1|}{\|[-\partial_x d, -\partial_y d, 1]\|}. \tag{16a}$$

The consistency loss penalizes misalignment between the predicted normals $n_{pred}$ and the approximated normals $n_{depth}$ as in (16b):

$$L_{x-dn} = 1 - \frac{n_{pred} \cdot n_{depth}}{\|n_{pred}\|\|n_{depth}\|}, \tag{16b}$$

This approach, inspired by [23], reinforces geometric consistency by aligning the predicted normals with those derived from the depth map.

**Edge-Segmentation Consistency Loss** computes the spatial gradient of the segmentation logits $S$ as $\nabla S = \left(\frac{\partial S}{\partial x}, \frac{\partial S}{\partial y}\right)$ and measures the absolute difference from the edge prediction $E$ using the $L_1$ norm as in (17),

$$L_{x-es} = \|\nabla S - E\|_1. \tag{17}$$

This approach enforces a tight correspondence between semantic boundaries and detected edges, drawing inspiration from TriangleNet [24].

### C. Task Balancing

We use Dynamic Weight Averaging (DWA) [25] to balance the learning pace across tasks. For each task $i$, the weight at step $t$ is computed as in (18).

$$w_i(t) = \frac{N \cdot \exp\left(\frac{r_i(t-1)}{T}\right)}{\sum_{n=1}^{N} \exp\left(\frac{r_n(t-1)}{T}\right)}, \tag{18}$$

$$\text{with } r_i(t-1) = \frac{L_i(t-1)}{L_i(t-2)}, \tag{18b}$$

where $N$ is the number of tasks, $L_i(t)$ is the task-specific loss, and $T$ controls the softness of the weighting. This approach enables efficient balancing of the learning dynamics without requiring additional backward passes to compute task-specific gradients as in GradNorm [7]. We apply DWA to the four task-specific losses ($L_{seg}, L_{depth}, L_{normals}, L_{edges}$ ) at all stages, with $T = 2$. The cross-task consistency losses ($L_{x-dn}, L_{x-es}$) are introduced only during fine-tuning with a fixed weight of 0.1, ensuring their magnitudes remain aligned with the single-task losses for a balanced overall objective.

## IV. EXPERIMENTS

Our network is designed to enhance indoor spatial mapping, and to validate its effectiveness, we evaluate our approach on diverse datasets.

### A. Datasets

We evaluate our approach on two indoor datasets, NYUDv2 [26] and Hypersim [27], and on a single outdoor dataset, Cityscapes [28]. NYUDv2 consists of 1,449 real-world indoor RGB-D images (640×480) with 40 semantic categories (795 for training, 654 for testing). Hypersim is a photorealistic synthetic dataset of 77,400 images (74,619 publicly available) from 461 scenes, following NYUDv2's labeling but omitting people in the public dataset, which can introduce class imbalance. Cityscapes captures urban street scenes in 50 German cities and provides around 5,000 annotated images (2,975 for training, 500 for validation, and 1,525 for testing) at a resolution of 2048×1024. Following previous work [12], we generate disparity and convert it to logarithmic depth to address the lack of dense depth annotations in Cityscapes.

TABLE I. PERFORMANCE COMPARISON OF M2H FRAMEWORK VERSUS SOTA METHODS ON NYUDV2 VALIDATION SET.

| Method | Semseg mIoU↑ | Depth RMSE↓ | Normal mErr↓ | Boundary odsF↑ |
|---|---|---|---|---|
| MTI-Net [14] | 45.97 | 0.5365 | 20.27 | 77.86 |
| ATRC [16] | 46.33 | 0.5363 | 20.18 | 77.94 |
| InvPT [18] | 53.56 | 0.5183 | 19.04 | 78.10 |
| TaskPrompter [29] | 55.30 | 0.5152 | 18.47 | 78.20 |
| MQTransformer [30] | 54.84 | 0.5325 | 19.67 | 78.20 |
| MTMamba [10] | 55.82 | 0.5066 | 18.63 | 78.70 |
| InvPt+Bi-MTPD-C [15] | 54.86 | 0.515 | 19.50 | 78.20 |
| MLoRE [17] | 55.96 | 0.5076 | 18.33 | 78.43 |
| MTMamba++ [11] | 57.01 | 0.4818 | 18.27 | 79.40 |
| SwinMTL [12] | 58.14 | 0.5179 | na | na |
| M2H-small | 58.05 | 0.4365 | 14.04 | 74.44 |
| M2H | **61.54** | **0.4196** | **13.81** | **85.27** |

(Note: ↑ indicates that a higher result corresponds to better performance, whereas ↓ indicates that a lower result is preferable.)

TABLE II. COMPARISON WITH SOTA SINGLE TASK METHODS ON HYPERSIM V1 SPLIT TEST SET

| Method | Semseg mIoU↑ | Depth RMSE↓ | AbsRel ↓ | δ<1.25 ↑ |
|---|---|---|---|---|
| ScaleDepth-NK [2] | na | 4.825 | 0.381 | 0.413 |
| EMSANet [3] | 46.66 | na | na | na |
| Depth Anything[a] [31] | na | na | 0.363 | 0.361 |
| ZoeDepth[a] [32] | na | 5.77 | 0.419 | 0.274 |
| M2H | **52.31** | **3.19** | **0.326** | **0.532** |

a. Depth anything and Zoe Depth results are zero-shot

TABLE III. PERFORMANCE COMPARISON OF M2H FRAMEWORK VERSUS SOTA METHODS ON CITYSCAPES VALIDATION SET

| Method | Semseg mIoU↑ | Depth RMSE↓ |
|---|---|---|
| MGNet [33] | 55.70 | 8.300 |
| PAD-Net [34] | 70.23 | 6.777 |
| 3-ways [35] | 75.00 | 6.528 |
| SwinMTL [11] | 76.41 | 6.32 |
| M2H | **77.6** | **6.10** |

## B. Experimental Setup

Our approach is designed to support any ViT-based encoder. In our experiments, we use DINOv2 small [12] due to its efficiency and rich feature representation compared to other ViT backbones, all packaged within an architecture that enables faster inference. We optimize the network using AdamW with a base learning rate of 5e-4 and a weight decay of 1e-4, following a polynomial learning rate schedule with a decay factor of 0.9. Data augmentations, including random adjustments to brightness, contrast, gamma, hue-saturation, and horizontal flipping, are applied following established methodologies [10], [12]. For direct comparisons, we set the maximum depth range to 80m for Cityscapes, 10m for NYUDv2, and 20m for Hypersim. Our implementation is trained on an A40 GPU and tested on an RTX3080 laptop GPU.

Following [18], we evaluate our approach using the mean intersection over union (mIoU) for semantic segmentation, root mean square error (RMSE) for depth estimation, mean error (mErr) for surface normal estimation, and the optimal-dataset-scale F-measure (odsF) for object boundary detection.

## C. Results

As compared in Table I, on NYUDv2, M2H leverages multi-task interactions to achieve superior performance in dense prediction tasks compared to existing methods.

TABLE IV. PARAMETER & GFLOPS ANALYSIS ON NYUDV2 DATASET

| Method | #P | GFLOPs |
|---|---|---|
| TaskPrompter [29] | 373M | 416 |
| SwinMTL[a] [12] | 87.38M | 65 |
| MTMamba++ [11] | 315M | 524 |
| M2H-small | **33.7M** | **59** |
| M2H | 81M | 488 |

Furthermore, as reported in Table II, on Hypersim, M2H outperforms zero-shot, task-specific models; ZOEdepth [32] and Depth Anything [31], which are trained on substantially larger datasets, underscoring the benefits of multi-task learning under data-constrained scenarios.

As shown in Table III, our M2H framework outperforms the current multi-tasking SOTA on the Cityscapes validation set, achieving a +1.2 mIoU over SwinMTL [12] and reducing depth RMSE by approximately 3.5%. Notably, even though M2H is primarily tailored for indoor scenarios with additional tasks, it effectively generalizes to outdoor environments, underscoring the robustness and flexibility of our multi-task approach. In addition, our distilled variant, M2H-Small, achieves competitive semantic segmentation performance and a substantial RMSE improvement over SOTA methods, all within a reduced computational budget, as in Table IV; it uses 64 channels instead of 256 in M2H, along with depth-wise convolutions, while retaining the same window patch size.

Finally, we evaluated our approach using Mono Hydra [4], a framework that generates 3D scene graphs for high-level visual perception tasks in autonomous agents by leveraging real-time depth and semantic predictions. For this evaluation, we employed the ITC dataset [4], which comprises image and IMU sequences. As shown in Table V, our experiments, conducted on a laptop GPU (RTX 3080) with a 224×224 input image sequence, demonstrate that our multi-task models outperform the combined single-task approach employed in Mono Hydra. We further compare our methods to MTMamba++ [11], the next best-performing model for depth prediction, as shown in Table I. Our approach achieves lower error metrics, including approximately a 2x improvement in mean error (ME) and a reduction in standard deviation (SD), as defined in Table V, all while maintaining real-time performance. For fairness, we used MTMamba++ [11] and M2H weights, both of which were trained on the same dataset, NYUDv2.

## D. Ablation results

Results in Table VI demonstrate the importance of both WMCA and GGFM modules. Replacing WMCA with multi-head attention (MHA) increases overhead but yields weaker performance, and removing WMCA entirely or omitting global context also degrades results across tasks, particularly for depth and semantic segmentation.

TABLE V. 3D MAPPING TEST WITH ITC DATASET BASED ON MONO HYDRA FRAMEWORK FOR MONOCULAR SPATIAL PERCEPTION

| Model | 2nd Floor | | 3rd Floor | | FPS |
| | ME (m)↓ | SD (m)↓ | ME (m)↓ | SD (m)↓ | |
|---|---|---|---|---|---|
| DistDepth[36]+HRNetv2[37] | 0.19 | 0.18 | 0.21 | 0.16 | 15 |
| MTMamba++ [11] | 0.21 | 0.22 | 0.18 | 0.19 | 18 |
| M2H-small | 0.16 | 0.18 | 0.15 | 0.17 | 42 |
| M2H | 0.11 | 0.14 | 0.10 | 0.13 | 30 |

TABLE VI. ABLATION STUDIES ON THE NYUDV2 DATASET

| Ablation | Semseg mIoU↑ | Depth RMSE↓ | Normal mErr↓ | Boundary odsF↑ | #p | FLOPs |
|---|---|---|---|---|---|---|
| MHA w/o WMCA | 50.94 | 0.5204 | 19.86 | 77.74 | 73M | 592 |
| w/o WMCA | 51.67 | 0.5202 | 18.17 | 72.45 | 38M | 213 |
| w/o GGFM | 56.92 | 0.5087 | 20.71 | 79.29 | 55M | 446 |

## V. CONCLUSION

In summary, we introduce a multi-task dense scene understanding framework that fuses depth, semantics, boundaries, and surface normals via a novel WMCA block for local feature sharing and a parallel global path for contextual aggregation. Our approach sets a new benchmark by outperforming current SOTA methods in both indoor and outdoor scenarios. Crucially, ensuring temporal consistency for downstream tasks like 3D mapping remains an important challenge. Future work will focus on refining temporal stability and expanding training to diverse datasets, paving the way for more resilient and adaptable real-world applications.